\documentclass[10pt,letterpaper,twocolumn]{article}
\usepackage[T1]{fontenc}
\usepackage{times}
\usepackage[letterpaper,textwidth=7in,textheight=9.1in,centering,footskip=26pt]{geometry}
\usepackage{amsmath,amssymb,graphicx,booktabs,array,textcomp}
\usepackage[table]{xcolor}
\usepackage{placeins}
\usepackage[font=small,labelfont=bf,labelsep=period]{caption}
\usepackage[numbers,sort&compress]{natbib}
\usepackage{titlesec}
\titleformat{\section}{\large\bfseries}{\thesection.}{.5em}{}
\titleformat{\subsection}{\normalsize\bfseries}{\thesubsection.}{.5em}{}
\titlespacing*{\section}{0pt}{2ex plus .5ex minus .3ex}{.8ex}
\titlespacing*{\subsection}{0pt}{1.7ex plus .5ex minus .3ex}{.6ex}
\definecolor{inkblue}{rgb}{0.15,0.34,0.49}
\usepackage[breaklinks,colorlinks,allcolors=inkblue]{hyperref}
\hypersetup{pdftitle={Agreement Is Not Accuracy: LiDAR Elevation Errors Hidden in LiDAR–Camera Extrinsics},
 pdfauthor={Yeon Kun Lee, Dongseok Yang, Deok-Kyeong Jang, Choongho Chung, Junyoung Lee, Hyeongseok So}}
\graphicspath{{figures/}}
\newcommand{\vect}[1]{\mathbf{#1}}
\newcommand{\degree}{^{\circ}}
\begin{document}
\twocolumn[{
\begin{center}
{\LARGE\bfseries Agreement Is Not Accuracy: LiDAR Elevation Errors\\[2pt]
Hidden in LiDAR--Camera Extrinsics\par}
\vspace{11pt}
{\large Yeon Kun Lee\qquad Dongseok Yang\qquad Deok-Kyeong Jang\\[3pt]
Choongho Chung\qquad Junyoung Lee\qquad Hyeongseok So\par}
\vspace{4pt}
{MOVIN\quad\href{https://movin3d.com}{\texttt{movin3d.com}}\par}
\end{center}
\vspace{4pt}
}]
\raggedbottom
\begin{abstract}
LiDAR and cameras provide complementary measurements for 3D vision, yet agreement between their observations does not establish the accuracy of either sensor's internal geometry. We show that systematic LiDAR elevation biases can be absorbed into the estimated extrinsics. Using synchronized LiDAR intensities and camera-estimated checkerboard poses, we jointly estimate a sensor-specific elevation correction shared across recordings and extrinsics for each recording. Ray--plane intersections locate the checker pattern without using reported ranges, preventing range errors from directly moving the lookup. Under matched intensity and correction models, this lookup gives smaller camera-consistency residuals than measured-point lookup. Across six identically designed rigs using the same sensor models, recording-excluded corrections reduce the mean translation-reference distance from 7.57 to 3.04\,mm. On separate structures, they reduce inter-plane angle errors against references from independent dense camera scans and improve image-edge alignment. Rolling the rigs shifts the plane pairs exhibiting the largest angular improvements, as predicted for an elevation-only correction. Within the ray--plane model, calibration-target projections and loop-consistency residuals change little despite these geometric improvements. These findings show why agreement after extrinsic calibration need not establish correct internal LiDAR geometry. Synchronized observations of a familiar checkerboard provide a practical route to estimating and testing elevation corrections without a dedicated large calibration field. We plan to release code and data for reproducing the main results.
\end{abstract}

\begin{figure*}[!t]
\centering
\includegraphics[width=.94\textwidth]{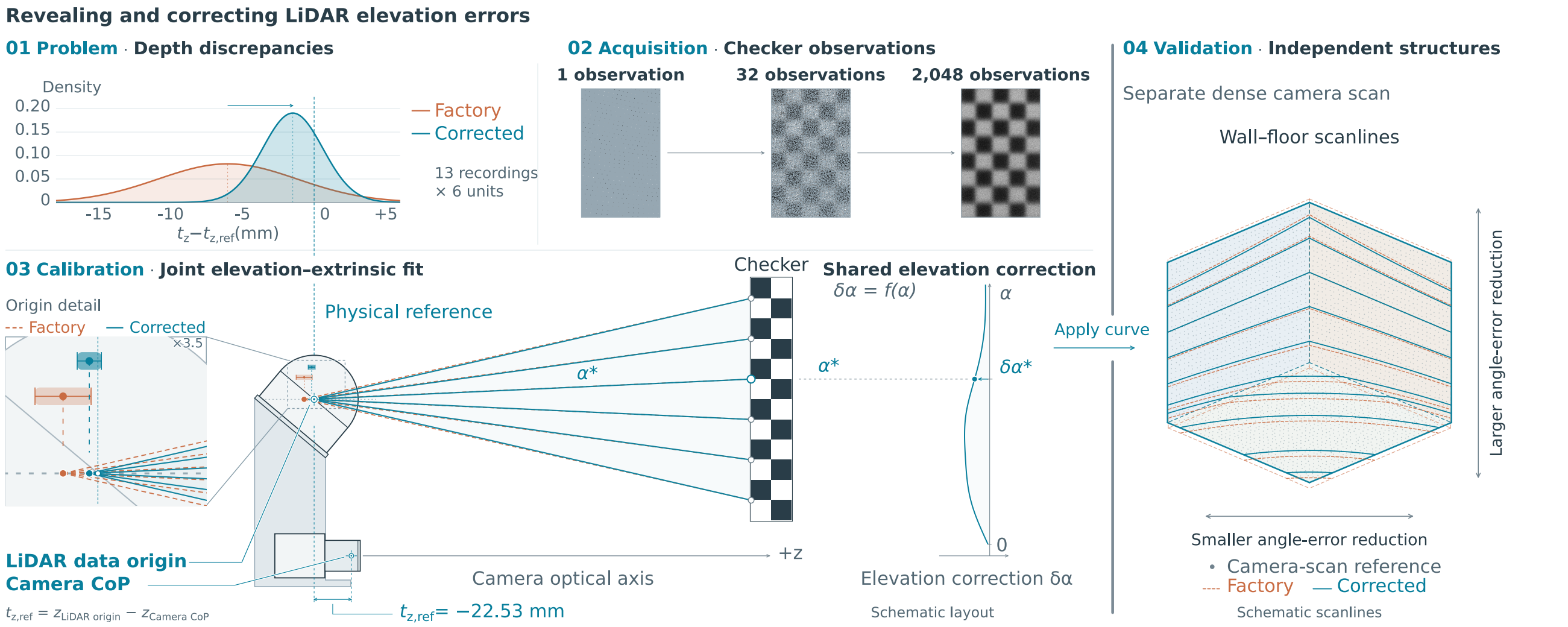}
\caption{Overview. (01) Gaussian summaries of signed depth-reference differences across 13 recordings $\times$ six units. (02) Synchronized checker returns. (03) Shared elevation and per-recording extrinsics; bands show the same means $\pm$ one standard deviation as (01), at a different depth scale. (04) Separate-structure evaluation. Fitting uses individual observation poses. Rig, rays and upright scanlines are schematic; the correction curve is estimated from one tested unit. Physical and camera-scan references are for evaluation.}
\label{fig:overview}
\end{figure*}

\section{Introduction}\label{sec:intro}

{\widowpenalty=10000
Estimating camera intrinsics is a standard calibration step in 3D vision, yet many LiDAR--camera calibration methods hold manufacturer-provided LiDAR geometry fixed. Residual internal errors can then be partly absorbed by the estimated rigid transformation, so good alignment alone does not establish accurate internal sensor models. Prior work calibrates range and beam directions using geometric targets, calibration fields, or positioning mechanisms~\cite{glennie2010,chen2012,hyyti2026}. We ask whether familiar checkerboard observations can also support the inspection and calibration of LiDAR direction geometry.\par}

Our investigation began with unit-dependent differences between estimated translations and a physical reference across rigs of identical design, most prominently along the camera optical axis. The rigs used the same camera, lens, and LiDAR models on precision CNC-machined structures. We first examined camera intrinsics, projection centers, and mechanical placement. The discrepancies remaining after these checks motivated us to revisit the LiDAR geometry that had been treated as fixed.

Camera-reconstructed checkerboard poses revealed range residuals that differed across units and varied nonlinearly with distance. Measured checker points can therefore mix range biases with direction errors. We align observed intensities with Fu et al.'s differentiable checker~\cite{fu2023}, using ray--plane intersections to locate pattern samples without reported range. However, excluding reported range from the lookup did not eliminate the unit-dependent translation discrepancies. We therefore examine the channel-to-ray mapping, analogous to the pixel-to-ray mapping in camera calibration, and jointly estimate a sensor-specific elevation correction shared across recordings and extrinsics for each recording.

We test whether these corrections transfer beyond the recordings used to estimate them (Fig.~\ref{fig:overview}). Recording exclusion, sensor exchange, changes of observing camera, and separate structures observed in two rig orientations test sensor-specific reuse and improvements in scene geometry. Physical reference dimensions are used only for evaluation. Our contributions are:
\begin{itemize}
\item \textbf{Physical-reference evaluation of millimeter-level translation accuracy.} Across six rigs and 13 recordings, recording-excluded corrections reduce the mean translation-reference distance from 7.57 to 3.04\,mm. The physical reference reveals unit-dependent offsets that image alignment and loop consistency alone do not expose.
\item \textbf{Checkerboard-based elevation calibration.} We estimate elevation corrections and recording-specific extrinsics from synchronized checkerboard observations, without using reported range for pattern lookup, and compare lookup geometries under the same intensity objective and correction model.
\item \textbf{Validation beyond the calibration target.} We demonstrate reuse across recordings and after sensor exchange, with validation against independently measured structures and image boundaries.
\end{itemize}

\section{Background}\label{sec:related}

\noindent\textbf{Reported translation accuracy.} On real sensor data, Koide et al. report a 10\,mm translation difference from sphere-target calibration using multiple data pairs~\cite{koide2023}. Fu et al. report 2--4\,mm between LiDAR-composed camera transforms and image-only stereo calibration~\cite{fu2023}. On real KITTI data with synthetic miscalibration, LCCNet reports 2.97\,mm mean per-axis absolute error after temporal filtering~\cite{lccnet2021}. In simulation, Beltr\'an et al. report 8.2\,mm mean Euclidean translation error for monocular/HDL-64 calibration with three target poses~\cite{beltran2022}.

\noindent\textbf{Evaluation references.} These references and statistics differ (Supplementary Section~S1). Target residuals~\cite{fastcalib2025}, dataset and registration-derived transforms~\cite{miaslcec2024,tlccalib2026,soac2024,unical2024,mamv2xcalib2025,lirecnet2026}, and camera--LiDAR loops~\cite{gentilini2025} assess different quantities. We use physical sensor-center references across six units and 13 recordings, with sensor exchange and independent structures, to investigate systematic LiDAR errors.

\noindent\textbf{Internal LiDAR calibration.} Prior work estimates per-laser range terms~\cite{glennie2010,glennie2016}, beam angles~\cite{muhammad2010,chen2012,hyyti2026}, parameters from surface crispness~\cite{levinson2014}, and general intrinsic models~\cite{huang2020global}. Joint intrinsic--extrinsic calibration has estimated range terms with elevation fixed~\cite{mirzaei2012}. Separate planar observations test calibrated geometry~\cite{chen2012,huang2020global}, but planes constrain points only along their normals~\cite{chow2013}. We estimate elevation from camera-observed checker coordinates and test transfer through plane residuals, relative orientations and image boundaries.

\noindent\textbf{Range-dependent geometry.} Range errors depend on distance and incidence angle~\cite{laconte2019}; single-photon measurements can also depend on photon count~\cite{rapp2020,gupta2019}. A flat checker can therefore produce a distorted point cloud. We evaluate Fu~et~al.'s differentiable pattern~\cite{fu2023} at ray--plane intersections, separating its pattern model from range-dependent lookup geometry.

\noindent\textbf{Errors outside a rigid sensor model.} Windshield refraction~\cite{verbiest2020}, projective LiDAR artifacts~\cite{zhu2024projective}, and solid-state glare~\cite{gump2026ghosts} also motivate examining sensor effects alongside extrinsic alignment.

\begin{figure*}[!t]
\centering
\includegraphics[width=.94\textwidth]{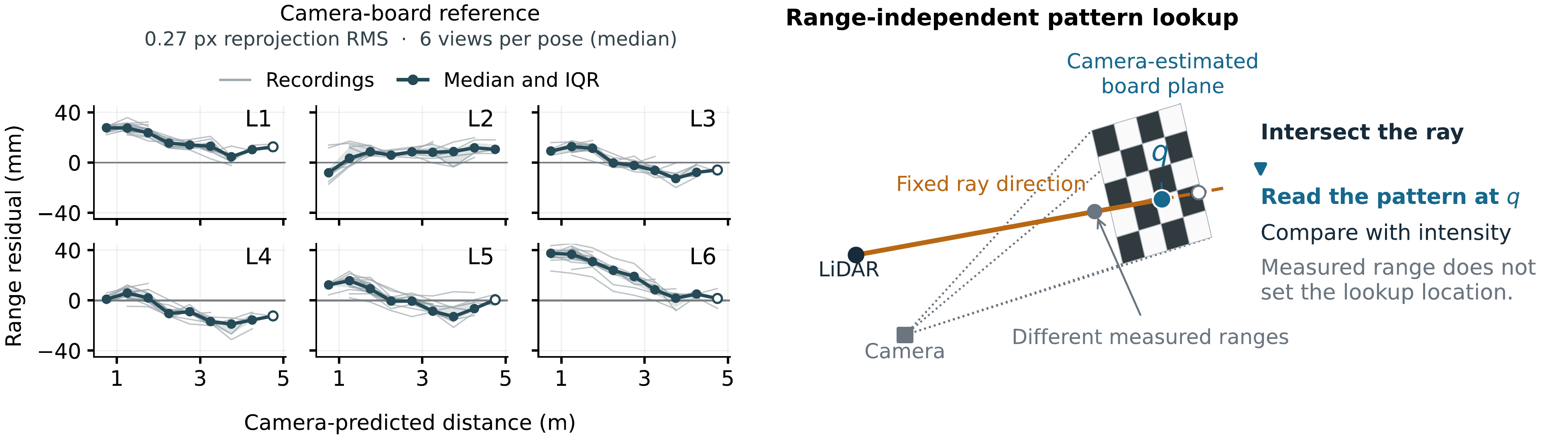}
\caption{Reported range and ray--plane lookup. Left: reported minus camera-predicted range after calibration (test observations, 0.5\,m bins). Gray: recording medians; dark line/shading: their median/IQR; open markers: fewer than three recordings. Right: changing range along a fixed ray leaves the board intersection $\vect{q}$ unchanged.}
\label{fig:ray_lookup}
\end{figure*}

\section{Checkerboard-Based Elevation Calibration}\label{sec:method}

We jointly estimate sensor-specific elevation corrections and per-recording extrinsics from synchronized LiDAR intensities and camera-estimated checkerboard poses. We determine checker-pattern lookup locations from ray--plane intersections, without using reported range.

\subsection{Synchronized checkerboard observations}\label{sec:sync}

A single LiDAR scan samples the checker sparsely. Moving the board through different positions and tilts provides complementary samples of the same known pattern. Synchronization associates each sparse sample with its own camera-estimated board pose, approximated as constant while the LiDAR sweeps the board. The sensors share a clock through the Precision Time Protocol (PTP), and a hardware trigger centers the exposure on the LiDAR rotor azimuth facing the camera. An \emph{observation} comprises one camera-estimated board pose and its LiDAR returns; a \emph{recording} is a sequence of observations. Slow board translation and rotation vary the sampled checker locations, distances, heights, and tilts (Fig.~\ref{fig:overview}, panel 02).

The calibration objective uses only the printed checker pattern. In our implementation, a retroreflective border initializes the extrinsics and checker phase; border returns are excluded from refinement. The target is a $600\times840$\,mm aluminum plate with 80\,mm checker squares and a 30\,mm border strip.

With our JT128 sensors rotating at 20\,Hz, the board returns span a median of 6.2\,ms per observation, comparable to the 7\,ms camera exposure. With the shared correction fixed, interpolating the board pose to each return's time changes the refitted extrinsics by at most 0.91\,mm and $0.048\degree$ across 78 unit--recording conditions (six units $\times$ 13 recordings).

\subsection{Pattern lookup on the ray and the board plane}\label{sec:lookup}

To avoid moving the checker lookup with a range error, we intersect the ray with the camera-reconstructed plane (Fig.~\ref{fig:ray_lookup}). Let $(R_{CB,k}, \vect{t}_{CB,k})$ be the board-to-camera transformation of observation $k$ and $(R, \vect{t})$ the LiDAR-to-camera transformation; the camera $+z$ axis is the optical axis, and $t_z$ is the depth of the LiDAR origin in camera coordinates. The board plane $z_B = 0$ is, in camera coordinates,
\begin{equation}
\vect{n}_k = R_{CB,k}\,\vect{e}_z,\quad d_k = -\vect{n}_k^{T}\vect{t}_{CB,k},\quad \vect{n}_k^{T}\vect{x}_C + d_k = 0 .\label{eq:plane}
\end{equation}
After vendor decoding, including the manufacturer's optical-origin correction, each XYZ return is normalized to a unit direction from the reported sensor origin; QT128 additionally uses the ray preprocessing based on its documented transmitter--receiver geometry~\cite{hesaiqt128}. This uses a central-ray approximation at the reported origin. The resulting $\vect{u}_i$ has elevation $\alpha_i$ and azimuth $\theta_i$ in the sensor's native frame, whose $z$ axis is the spin axis. With $\vect{u}'_i$ the corrected direction (Eq.~\eqref{eq:udir}), the board coordinates of the intersection are
\begin{align}
\lambda_{ki} &= -\frac{d_k + \vect{n}_k^{T}\vect{t}}{\vect{n}_k^{T} R\,\vect{u}'_i},\label{eq:lambda}\\
\vect{q}_{ki} &= \Pi_{xy} R_{CB,k}^{T}\left(\vect{t} + \lambda_{ki} R\,\vect{u}'_i - \vect{t}_{CB,k}\right),\label{eq:q}
\end{align}
where $\Pi_{xy}$ keeps the in-plane coordinates. The measured range $r_i$ does not appear. For fixed ray directions, origins, and selected returns, reported-range errors, including nonlinear ones, do not move the lookup position.

We use Fu~et~al.'s differentiable sinusoidal checker~\cite{fu2023} to define the virtual pattern in our board-coordinate phase, evaluated at the ray--plane intersections:
\begin{equation}
m(\vect{q}) = \sin\!\left(\frac{\pi x}{w}\right)\sin\!\left(\frac{\pi y}{w}\right),\qquad w = 80\,\mathrm{mm}.\label{eq:pattern}
\end{equation}
The virtual pattern specifies spatial variation but not the absolute scale of LiDAR intensity. We allow an observation-specific gain and offset. Eliminating these parameters from a least-squares mismatch and normalizing by the observed intensity variance gives $1-\rho_k^2$, where $\rho_k$ is the Pearson correlation between the virtual pattern and the measured intensities of observation $k$; we average across observations:
\begin{equation}
J(R,\vect{t},\vect{c}) = \frac{1}{K}\sum_{k=1}^{K}\left(1-\rho_k^{2}\right).\label{eq:cost}
\end{equation}
Here, $\vect{c}$ denotes the elevation-correction coefficients. Plane residuals and the physical reference are excluded from fitting.
The reflective border initializes the checker phase; final alignments are visually checked for phase slips. Intensities are percentile-normalized per channel and observation, and the selected inner-checker returns stay fixed during refinement. Normalization and support settings are given in Supplementary Section~S2.

\subsection{Joint elevation and extrinsic estimation}\label{sec:shared}

Each unit $u$ shares one elevation correction $f_u(\alpha;\vect{c}_u)$ across recordings spanning azimuth, elevation, and depth. Separate extrinsics $(R_{us},\vect{t}_{us})$ accommodate re-mounting between recordings $s$. The curve corrects elevation relative to the decoded vendor geometry. The correction changes a channel's cone opening angle; the extrinsic rotation reorients the cone as a whole. Azimuth coverage helps distinguish elevation offset from rigid rotation; depth diversity reduces angular--translation coupling, and elevation coverage constrains curve shape. Simulations and local Jacobian analysis of all fitting observations support these roles (Supplementary Section~S3). In the sensor's own frame only the elevation $\alpha$ is changed:
\begin{equation}
\vect{u}'_i =
\begin{bmatrix}
\cos(\alpha_i + f_u(\alpha_i;\vect{c}_u))\cos\theta_i\\
\cos(\alpha_i + f_u(\alpha_i;\vect{c}_u))\sin\theta_i\\
\sin(\alpha_i + f_u(\alpha_i;\vect{c}_u))
\end{bmatrix},\label{eq:udir}
\end{equation}
\begin{equation}
\min_{\vect{c}_u,\{R_{us},\vect{t}_{us}\}}
\frac{1}{|\mathcal{S}_u|}\sum_{s\in\mathcal{S}_u} J_{us}(R_{us},\vect{t}_{us},\vect{c}_u).\label{eq:joint}
\end{equation}
Recordings are weighted equally. A smooth natural cubic spline represents sensor-specific elevation bias without requiring internal optical design information. LiDAR transmit/receive optics and wide-field angular models~\cite{kannala2006} motivate radial-polynomial controls with fewer coefficients (Supplementary Section~S8).

Curve support requires sufficient fitting observations from at least two recordings. Reuse evaluation excludes the evaluated recording from curve and support estimation, fits only its extrinsics with the curve fixed, and evaluates its held-out observations.

\subsection{Properties of the ray--plane lookup}\label{sec:properties}

For fixed directions, origins, selected returns and board poses, reported-range perturbations leave the ray--plane lookup unchanged. Range still enters preprocessing and selection. Section~\ref{sec:benchmark} compares this property with measured-point lookup~\cite{fu2023}.

A distinct ambiguity affects multi-camera consistency. If two LiDAR-to-camera solutions share a right-multiplicative error $E$, it cancels in their relative transform:
\[
(T_{jL}E)(T_{iL}E)^{-1}=T_{jL}T_{iL}^{-1}.
\]
Thus a small loop residual constrains the difference between the errors, not their common component. Section~\ref{sec:blind} tests this using independently fitted camera--LiDAR transforms.

\section{Calibration Discrepancies and Consistency}\label{sec:hides}

\subsection{Acquisition and physical reference}\label{sec:recordings}

Six aluminum rigs share the same design, CNC-machined to approximately 0.1\,mm precision. Each carries a FLIR Blackfly S (BFS) global-shutter camera (1.6\,MP) and a JT128 LiDAR in the same configuration. We label the six main cameras C1--C6; two auxiliary wide-angle cameras, wideC1 and wideC2, provide additional board views. Board poses are obtained by multi-view bundle adjustment with intrinsics fixed from separate camera-calibration recordings. The poses are expressed in each rig's main-camera frame and remain fixed, together with camera geometry, during LiDAR fitting.

The main dataset comprises 13 recordings of the same six units. To examine camera and mounting contributions, we varied board distance, motion coverage, and rig arrangement, and repeated observations after re-mounting. A subsequent LiDAR exchange with cameras fixed tests whether discrepancies follow the rigs or the sensors. Data were acquired on five dates within three weeks, with at least one hour of sensor warm-up before each acquisition. The valid data used for analysis comprise approximately 94,000 unit-observations and 90 million returns. We reserve observations at fixed timestamps for evaluation and use the remaining valid observations within the supported elevation interval for fitting. Comparisons between factory and corrected directions use the same camera geometry and identical sets of fitting and held-out observations.

Factory angle corrections and optical-origin decoding report points in the manufacturer-specified LiDAR frame. Downstream extrinsic calibration estimates the pose of this frame. We therefore use its documented origin as a fixed physical reference: the rig CNC drawing, the JT128 origin on its rotation axis~\cite{hesaijt128}, and the camera center of projection give $\mathbf{t}_{\mathrm{ref}}=[0,-94.76,-22.53]$\,mm. Our primary measure is the mean translation-reference distance, $\operatorname{mean}\|\Delta\mathbf{t}\|$, where $\Delta\mathbf{t}=\mathbf{t}-\mathbf{t}_{\mathrm{ref}}$. Before correction, the optical-axis component accounts for much of the translation discrepancy (Table~\ref{tab:phenomenon}). Figure~\ref{fig:overview}, panel 01, shows this signed depth component, $t_z-t_{z,\mathrm{ref}}$, across units and recordings. Table~\ref{tab:budget} summarizes camera and mounting consistency controls. We interpret physical-reference differences alongside these controls, sensor exchange, and transfer to separate structures.

\begin{table}[t]
\caption{Camera and mounting consistency controls, and the LiDAR data-origin reference.}
\label{tab:budget}
\centering
\footnotesize
\setlength{\tabcolsep}{3pt}
\begin{tabular}{p{0.60\columnwidth}r}
\toprule
Control or reference & Quantity or definition \\
\midrule
Coplanar camera-front acquisitions & 0.13/0.17\,mm RMS \\
Camera-chain repeatability, five recordings & 0.02--0.06\,mm \\
Bracket mount height agreement & 0.2--0.6\,mm \\
Focal-length perturbation, effect on $t_z$ & $\le0.55$\,mm \\
LiDAR re-seating, translation change & $\le1.7$\,mm \\
Arc-fixture camera-center consistency & 0.11\,mm RMS \\
JT128 data-origin position & Manufacturer-specified \\
\bottomrule
\end{tabular}
\end{table}

We measured inter-plane angles using separate camera scans of the two structures. A separate unit of the same FLIR camera model was used, with intrinsics fixed from a dedicated calibration session. COLMAP \cite{schoenberger2016sfm,schoenberger2016mvs} reconstructed the surfaces at approximately 500,000 points/m$^2$. Table~\ref{tab:structure_references} reports the measured angles and their departures from $90^\circ$. These references are not used in LiDAR calibration fitting.

\begin{table}[t]
\caption{Separately measured structure angles (degrees). The $\pm$ values denote one standard deviation across 200 spatial block-bootstrap replicates using 25\,cm cells; $\Delta_{90}$ is the signed departure from $90^\circ$.}
\label{tab:structure_references}
\centering\footnotesize
\setlength{\tabcolsep}{4pt}
\begin{tabular}{llrr}\toprule
Structure & Plane pair & Measured angle & $\Delta_{90}$ \\
\midrule
1 & Left--right wall & $90.017\pm0.016$ & $+0.017$ \\
1 & Left wall--floor & $89.982\pm0.020$ & $-0.018$ \\
1 & Right wall--floor & $90.023\pm0.015$ & $+0.023$ \\
\midrule
2 & Left--right wall & $89.993\pm0.019$ & $-0.007$ \\
2 & Left wall--floor & $89.983\pm0.018$ & $-0.017$ \\
2 & Right wall--floor & $90.098\pm0.017$ & $+0.098$ \\
\bottomrule
\end{tabular}
\end{table}

\subsection{Unit-specific translation discrepancies}\label{sec:spread}

We first estimate only the per-recording extrinsics $R,\mathbf{t}$ using Eq.~\eqref{eq:cost}, without applying an additional elevation correction. The mean translation-reference distance is 7.57\,mm (Table~\ref{tab:phenomenon}). The largest differences occur along the optical axis: unit-mean signed depth differences range from $-11.56$ to $+2.30$\,mm (Fig.~\ref{fig:overview}, panel 01). Within-unit depth SDs are 1.41--3.09\,mm, including re-mounting. Thus, despite the identical rig design, the estimated translations show recurring unit-specific discrepancies.

This fitting already excludes reported range from checker-pattern lookup. The remaining translation discrepancies show that removing this direct range dependence alone does not resolve them. They motivate examining the LiDAR direction mapping as a calibration variable alongside the extrinsics.

\begin{table}[t]
\caption{Translation differences from the physical reference (mm). Top: signed depth-reference difference means and recording SDs over 13 recordings per unit; corrections are learned from the other recordings. Bottom: SD$_u$ is the SD of the unit mean depths; $|\Delta x|$, $|\Delta y|$ and $|\Delta z|$ are mean absolute component differences, and the translation-reference distance is $\operatorname{mean}\|\mathbf{t}-\mathbf{t}_{\mathrm{ref}}\|$, over all 78 conditions. Brackets give crossed-bootstrap 95\% intervals.}
\label{tab:phenomenon}
\centering
\footnotesize
\setlength{\tabcolsep}{4pt}
\resizebox{\columnwidth}{!}{%
\begin{tabular}{lrrrr}
\toprule
\multicolumn{5}{c}{Depth difference $t_z-t_{z,\mathrm{ref}}$ (mm)} \\
 & \multicolumn{2}{c}{Factory directions} & \multicolumn{2}{c}{With correction} \\
LiDAR unit & Mean & Rec.\ SD & Mean & Rec.\ SD \\
\midrule
L1 & $-5.78$ & 1.41 & $-3.31$ & 1.06 \\
L2 & $-11.56$ & 3.09 & $-2.04$ & 1.26 \\
L3 & $-9.03$ & 2.30 & $-2.60$ & 0.86 \\
L4 & $-5.35$ & 2.43 & $+0.51$ & 1.08 \\
L5 & $-6.77$ & 2.66 & $-2.48$ & 1.99 \\
L6 & $+2.30$ & 1.43 & $+0.98$ & 1.52 \\
\midrule
Between-unit SD$_u$ & \multicolumn{2}{r}{4.69} & \multicolumn{2}{r}{1.79} \\
\midrule
\multicolumn{5}{l}{Mean reference differences over all 78 conditions} \\
$|\Delta x|$ & \multicolumn{2}{r}{1.48} & \multicolumn{2}{r}{1.17} \\
$|\Delta y|$ & \multicolumn{2}{r}{2.38} & \multicolumn{2}{r}{1.29} \\
$|\Delta z|$ (depth) & \multicolumn{2}{r}{6.80} & \multicolumn{2}{r}{2.13} \\
$\|\Delta\vect{t}\|$ [95\%] & \multicolumn{2}{r}{7.57 [5.0, 10.5]} & \multicolumn{2}{r}{3.04 [2.2, 3.9]} \\
\bottomrule
\end{tabular}}
\end{table}

\subsection{The correction follows the LiDAR}\label{sec:exchange}

L2 and L6 exchanged rigs while the cameras stayed in place. Curves learned from the ten recordings in the original mounting configuration were held fixed and applied to three recordings after the exchange. Only the extrinsics were re-estimated for each post-exchange recording.

After the exchange, translation-reference distances decreased when each LiDAR retained its own correction curve. Applying the curve of the LiDAR previously mounted on that rig increased the unit-mean distances. Each LiDAR's own curve yielded a smaller distance than either alternative in all six unit--recording conditions (Table~\ref{tab:exchange}). Together with the re-mounting control, these results support an association between the reusable correction and the LiDAR unit rather than the camera or mounting position.

\begin{table}[t]
\centering\footnotesize
\caption{Translation-reference distances after sensor exchange (mm): mean $\|\mathbf{t}-\mathbf{t}_{\mathrm{ref}}\|$ over three recordings.}
\label{tab:exchange}
\setlength{\tabcolsep}{4pt}
\begin{tabular}{@{}lrr@{}}\toprule
Correction applied & L2 & L6 \\
\midrule
No additional elevation correction & 8.70 & 1.85 \\
Each LiDAR's own curve & \textbf{2.14} & \textbf{1.24} \\
Curve of the previous LiDAR on the rig & 9.48 & 8.69 \\
\bottomrule
\end{tabular}
\end{table}

\subsection{Calibration constraints and camera consistency}\label{sec:benchmark}

With consistent sensor geometry and accurate target observations, plane, boundary and pattern constraints should support the same physical transformation. Yet they yield different extrinsics from the same recordings and fixed sensor arrangement.

\begin{table*}[t]
\centering
\caption{Calibration with factory directions (No) or shared elevation corrections (Yes). Translation reports mean distances and uses 78 conditions and recording-excluded curves for corrected fits; the other-camera metric uses the paired protocol in S6. Full-data curves evaluate 198 planes, 180 pairs and 60 image segments. Proj.: held-out projection displacement from each method's factory projections, not reprojection error. Loop: mean over 23 common loops. Range $+30$: translation change after $+30$\,mm, with directions, curves and selection fixed. Ref.: factory condition. $\dagger$ 381 camera pairs; otherwise 387. $\ddagger$ Some source fits reach bounds (S6).}
\label{tab:benchmark}
\footnotesize
\setlength{\tabcolsep}{3pt}
\renewcommand{\arraystretch}{1.10}
\resizebox{\textwidth}{!}{%
\begin{tabular}{llrrrrrrrrr}
\toprule
 & & \multicolumn{2}{c}{Translation} & \multicolumn{3}{c}{Scene geometry} & \multicolumn{2}{c}{Image} & \multicolumn{2}{c}{Consistency / sensitivity} \\
\cmidrule(lr){3-4}\cmidrule(lr){5-7}\cmidrule(lr){8-9}\cmidrule(lr){10-11}
Method & Elev. & $\|\Delta\mathbf t\|\downarrow$ & $\|\Delta\mathbf t_{\mathrm{other}}\|\downarrow$ & Plane RMS $\downarrow$ & Angle MAE $\downarrow$ & Improved $\uparrow$ & MAE $\downarrow$ & Proj. & Loop $\downarrow$ & Range $+30\downarrow$ \\
 &  & (mm) & (mm) & (mm) & ($^\circ$) & (of 180) & (px) & (px) & (mm) & (mm) \\
\midrule
Line + plane~\cite{zhou2018} & No & 16.89 & 19.45$^{\dagger}$ & 6.734 & 0.453 & ref. & 6.85 & --- & 2.06 & 27.99 \\
\addlinespace
Point-to-plane~\cite{unnikrishnan2005} & No & 28.63 & 33.66$^{\ddagger}$ & 6.734 & 0.453 & ref. & 4.91 & ref. & 5.00 & 31.22 \\
Point-to-plane~\cite{unnikrishnan2005} & Yes & 27.75 & 32.22$^{\ddagger}$ & 7.404 & 0.405 & 100 & 2.63 & 3.87 & 4.99 & 31.19 \\
\addlinespace
Fu-style + SSD~\cite{fu2023} & No & 7.15 & 14.24 & 6.734 & 0.453 & ref. & 4.08 & ref. & 8.71 & 4.14 \\
Fu-style + SSD~\cite{fu2023} & Yes & 3.43 & 11.71$^{\ddagger}$ & 6.724 & 0.346 & 150 & 2.61 & 0.98 & 9.33 & 4.18 \\
\addlinespace
Fu-style + squared Pearson & No & 8.16 & 14.80 & 6.734 & 0.453 & ref. & 4.07 & ref. & 7.27 & 3.59 \\
Fu-style + squared Pearson & Yes & 3.53 & 10.49 & 6.707 & 0.342 & 152 & 2.55 & 1.03 & 8.03 & 3.64 \\
\addlinespace
Ray + squared Pearson & No & 7.57 & 8.99 & 6.734 & 0.453 & ref. & 4.17 & ref. & 1.10 & 0.00 \\
Ray + squared Pearson & Yes & 3.04 & 3.23 & 6.713 & 0.343 & 151 & 2.73 & 0.96 & 1.10 & 0.00 \\
\bottomrule
\end{tabular}%
}
\end{table*}

Table~\ref{tab:benchmark} compares point-to-plane~\cite{unnikrishnan2005}, line-and-plane~\cite{zhou2018}, and Fu-style measured-point checker fitting with SSD~\cite{fu2023}. We isolate lookup geometry by also comparing measured-point and ray--plane lookup with the same squared Pearson objective and seven-knot elevation model. Factory fits estimate only extrinsics; joint fits add a shared elevation correction. No range correction is fitted (Supplementary Section~S5). Translation columns measure differences from the physical reference. $\|\Delta\mathbf{t}_{\mathrm{other}}\|$ uses the same reference after refitting extrinsics with another observing camera (Section~\ref{sec:camera_transfer}).

All three checker-pattern fits reduce translation-reference distances and structure-angle errors. Their mean own-camera translation-reference distances are similar. The matched Pearson comparison separates more clearly through camera consistency: mean loop residuals against an image-only reference are 1.10/8.03\,mm for ray--plane/measured-point lookup over 23 common loops, with smaller ray--plane residuals for both camera pairs.

Measured-point lookup projects from the camera center onto the board; range-displaced points can therefore indicate different pattern locations across cameras. Ray--plane lookup removes this dependence (Section~\ref{sec:properties}). In factory-direction fits, a 30\,mm range perturbation produces mean translation changes of 3.59\,mm for measured-point Pearson, about 28--31\,mm for plane-based constraints, and zero for ray--plane lookup. This supports separating reported range from elevation fitting.

\subsection{What conventional calibration checks miss}\label{sec:blind}

Loop consistency can reveal camera-dependent discrepancies, but shared errors cancel in the relative transform (Section~\ref{sec:properties}). Within the ray--plane model, image--point-cloud agreement and loop consistency remain nearly unchanged after elevation correction.

\noindent\emph{Point-projection displacement:} Joint changes in rotation and translation can partly compensate for corrected ray directions, keeping point projections nearly unchanged (Fig.~\ref{fig:overview}, panel 03). Across all 78 conditions, translation changes by a mean of 5.3\,mm, while projections of the same held-out returns move by about 1\,px on average.

\noindent\emph{Loop consistency:} Independently fitted LiDAR-to-camera transforms are composed and compared with an image-only reference. Across the same 23 loops, elevation correction changes translation by a median of 6.5\,mm, while the median absolute paired change in loop residual is 0.003\,mm.

\section{Validation of Elevation Calibration}\label{sec:paired}

\subsection{Cross-recording transfer}\label{sec:transfer}

\begin{figure}[!t]
\centering
\includegraphics[width=.90\columnwidth]{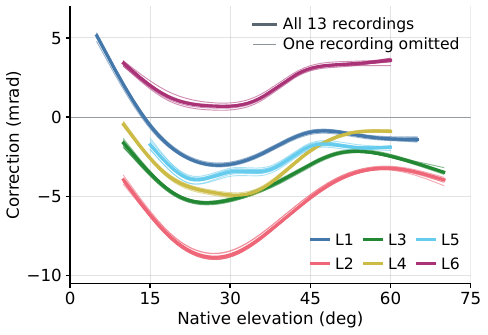}
\caption{Unit-specific elevation corrections learned across recordings. Thick curves use all 13 recordings over the supported interval; thin curves of the same color omit one recording at a time. The displayed curves stay within 0.42\,mrad of the full curve over their common plotted support. 1\,mrad $=0.057\degree$, i.e.\ 3\,mm of ray displacement at 3\,m.}
\label{fig:curves}
\end{figure}

\begin{table*}[t]
\definecolor{transferbg}{HTML}{DCEFEA}
\definecolor{transferfg}{HTML}{126B5F}
\centering
\caption{Mean absolute inter-plane angle error relative to the measurements in Table~\ref{tab:structure_references}, before and after checker-trained elevation correction (degrees). $\Delta$ is before minus after. Each configuration uses six JT128 units and four recordings; improved counts the pairs with smaller error. Planes are fitted independently to identical retained points without an orthogonality constraint (180 comparisons). Teal highlights wall--floor relations for upright rigs and wall--wall for rolled rigs.}
\label{tab:structures}
\small
\setlength{\tabcolsep}{3.2pt}
\renewcommand{\arraystretch}{1.15}
\begin{tabular}{lccc ccc ccc}
\toprule
 & \multicolumn{6}{c}{Orthogonal structure 1} & \multicolumn{3}{c}{Orthogonal structure 2} \\
\cmidrule(lr){2-7}\cmidrule(lr){8-10}
 & \multicolumn{3}{c}{Upright} & \multicolumn{3}{c}{Rolled $\approx90^\circ$} & \multicolumn{3}{c}{Upright} \\
\cmidrule(lr){2-4}\cmidrule(lr){5-7}\cmidrule(lr){8-10}
Plane pair & Before $\to$ after & $\Delta$ & Improved & Before $\to$ after & $\Delta$ & Improved & Before $\to$ after & $\Delta$ & Improved \\
\midrule
Left--right wall & $0.462\to0.448$ & $0.015$ & 14/18 & $0.590\to0.455$ & \cellcolor{transferbg}\textcolor{transferfg}{$\mathbf{0.134}$} & 18/18 & $0.203\to0.152$ & $0.051$ & 15/18 \\
Left wall--floor & $0.834\to0.543$ & \cellcolor{transferbg}\textcolor{transferfg}{$\mathbf{0.291}$} & 18/18 & $0.485\to0.461$ & $0.024$ & 15/18 & $0.295\to0.205$ & \cellcolor{transferbg}\textcolor{transferfg}{$\mathbf{0.090}$} & 20/24 \\
Right wall--floor & $0.576\to0.340$ & \cellcolor{transferbg}\textcolor{transferfg}{$\mathbf{0.236}$} & 24/24 & $0.364\to0.362$ & $0.002$ & 11/24 & $0.313\to0.162$ & \cellcolor{transferbg}\textcolor{transferfg}{$\mathbf{0.150}$} & 16/18 \\
\bottomrule
\end{tabular}
\end{table*}

\begin{figure*}[t]
\centering
\includegraphics[width=.90\textwidth]{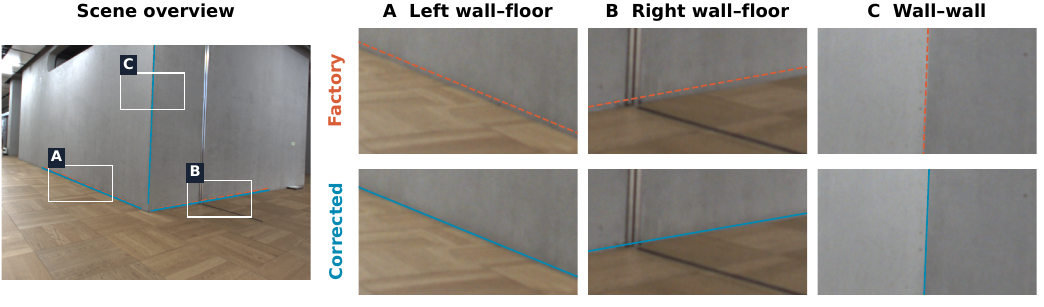}
\caption{Image alignment on an external structure (JT128). Orange (Factory) and blue (Corrected) mark fitted plane intersections using the corresponding extrinsics. Both wall--floor boundaries and the wall--wall boundary are shown. The example illustrates a large wall--floor improvement; the text reports all comparisons.}
\label{fig:jt_transfer}
\end{figure*}

Elevation corrections transfer to excluded recordings. We jointly fit the shared curve and extrinsics on the other recordings, then fix the curve and fit only the excluded recording's extrinsics. Neither fit uses the physical reference, yet translation-reference distances decrease in 76 of 78 conditions, with their mean falling from 7.57 to 3.04\,mm (Table~\ref{tab:phenomenon}). Held-out test costs also decrease in all 78 conditions. The recording-excluded curves show recurring shapes and levels (Fig.~\ref{fig:curves}). Excluding all recordings of one acquisition date at a time also yields a mean held-out translation-reference distance of 3.01\,mm (Supplementary Section~S6).

Across units, the curves share similar shapes but differ in level. We define level as the mean correction over the common $15$--$60^{\circ}$ interval and shape as the remaining variation. Pairwise shape correlations range from 0.87 to 0.99, whereas mean corrections range from $-6.3$ to $+2.0$\,mrad across units.

\subsection{Transfer across observing cameras}\label{sec:camera_transfer}

Replacing multi-view board poses with independent single-camera PnP estimates on matched observations changes the ray--plane curves by a mean RMS of 0.16\,mrad across six units. Similar elevation corrections can thus be estimated from single-camera poses.

Changing the observing camera gives a mean RMS curve difference of 0.24\,mrad. We then fix the own-camera curve, refit extrinsics with another camera, and transport them through the fixed camera chain to compare with the same physical reference. On paired observations, mean ray--plane reference differences are 3.08\,mm with the own camera and 3.23\,mm with another camera. The matched measured-point Pearson control gives 3.92 and 10.49\,mm.

The lookups give similar own-camera discrepancies but separate across observing cameras. This agrees with measured-point lookup geometry: wider baselines can amplify range-induced pattern displacement, whereas ray--plane lookup removes that dependence. Other-camera curves also retain improvement on the separate structures. Supplementary Section~S6 gives the paired protocol and results.

\subsection{Transfer to separate structures}\label{sec:structures}

Inter-plane angles, plane RMS and image-edge MAE probe surface orientation, planarity and alignment. A smooth correction may rotate a planar patch with little RMS change; image MAE also depends on model-matched extrinsics. We therefore report these complementary measurements together.

We apply fixed checker-trained curves to two structures, retaining reported ranges. The structure evaluation also includes more distant surfaces (median distance approximately $2.6\times$ that of the checker samples). The structures supply no calibration parameters. Six units provide 198 independently fitted planes and 180 pairs, evaluated against Table~\ref{tab:structure_references}. Rolling the rigs by approximately $90^{\circ}$ tests whether the dominant improvement follows the sensor's elevation direction.

The dominant angular improvement follows the rig rotation. Wall--floor relations improve most with upright rigs, whereas wall--wall relations improve most after rolling (Table~\ref{tab:structures}). This matches the predicted change in which plane pairs benefit from elevation correction. Of these upright wall--floor and rolled wall--wall pairs, 96 of 102 improve. Inter-plane angles are invariant to rigid extrinsics, so this directional change in the improvement supports transfer of the checker-trained correction to scene geometry.

Across both structures, 151 of 180 pairs improve, and mean absolute angle error falls from $0.453^{\circ}$ to $0.343^{\circ}$. Using $90^{\circ}$ as the reference gives the same overall trend, with 150 of 180 pairs improving. The other plane-pair orientations show smaller, mixed changes. Mean plane RMS remains nearly unchanged, from 6.73 to 6.71\,mm. Translation-reference and structural improvements also persist with alternative spline and radial parameterizations (Supplementary Section~S8).

Image alignment also improves at wall--floor boundaries when the elevation-corrected geometry is projected with its corresponding extrinsics (Fig.~\ref{fig:jt_transfer}). On one structure observed in four recordings, we measure the mean absolute perpendicular distance from projected plane intersections to detected image lines. Wall--floor MAE decreases from 5.31 to 3.22\,px, with all 42 comparisons improving. Wall--wall changes are mixed: seven of 18 comparisons improve, while mean MAE changes from 1.51 to 1.58\,px. These measurements test corrected geometry and extrinsics together. Supplementary Section~S7 gives evaluation details, detector sensitivity, and segment RMS.

\begin{figure}[!t]
\centering
\includegraphics[width=.95\columnwidth]{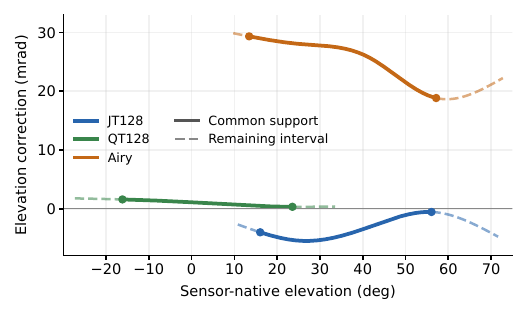}
\caption{Shared elevation corrections for additional units, learned from four recordings. Solid segments mark common observation support; dashed segments show the rest of the fitted interval. Positive values increase native elevation. This JT128 is separate from the six-unit study.}
\label{fig:additional_curves}
\end{figure}

\begin{table}[!t]
\caption{Transfer to a separate structure for one QT128 and one Airy unit. Angles use the structure 2 references in Table~\ref{tab:structure_references}. Each sensor contributes 18 plane pairs and 18 surfaces from six structure recordings.}
\label{tab:additional}
\centering\footnotesize
\setlength{\tabcolsep}{4pt}
\begin{tabular}{lcc}
\toprule
Quantity (Factory $\to$ Corrected) & QT128 & Airy \\
\midrule
Mean angle error ($\degree$) & $0.195\to0.173$ & $0.894\to0.359$ \\
Plane pairs improved & 14/18 & 18/18 \\
Mean plane RMS (mm) & $5.79\to5.77$ & $15.38\to15.09$ \\
Image MAE (px), wall--floor & $1.23\to0.98$ & $5.22\to1.56$ \\
Image MAE (px), wall--wall & $2.93\to2.95$ & $1.87\to1.19$ \\
Overall image MAE (px) & $1.79\to1.64$ & $4.10\to1.44$ \\
\bottomrule
\end{tabular}
\end{table}

\subsection{Other sensors}\label{sec:sensors}

We test one additional JT128, one QT128, and one Airy unit with five-knot curves from four checker recordings, evaluating each recording with the curve learned from the other three. Figure~\ref{fig:additional_curves} compares their shared curves on the same angular scale. Estimated elevation corrections range from $-5.5$ to $-0.5$\,mrad for the JT128, $+0.3$ to $+1.6$\,mrad for the QT128, and $+18.9$ to $+29.4$\,mrad for the Airy. The additional JT128 shares the main units' curve shape (correlations 0.89--1.00). The Airy correction remains large after verifying that the manufacturer's origin correction and unit-specific elevation table were applied~\cite{robosenseairy} (Supplementary Section~S7). Its recording-excluded curves differ from the full curve by less than 0.4\,mrad within common support.

\begin{samepage}
Separate structure measurements connect these fitted values to their geometric effects (Table~\ref{tab:additional}). Across six structure recordings, all 18 Airy plane-angle comparisons improve, alongside a smaller change in plane RMS. Its mean image-boundary MAE decreases from 4.10 to 1.44\,px, with larger reductions at wall--floor boundaries. QT128 has smaller angular changes and a small decrease in mean image-boundary MAE, from 1.79 to 1.64\,px. These results extend the evidence beyond JT128, while showing that the magnitude and geometric effect of the correction vary across the tested sensors.
\end{samepage}

\raggedbottom
\begin{samepage}
\section{Conclusion}\label{sec:conclusion}

Joint checkerboard fitting reduces translation-reference distances and yields sensor-specific elevation corrections that transfer across recordings, cameras and separate structures. Sensor exchange and rig rotation connect the improvements to the sensor and its elevation direction. Camera consistency and range sensitivity support ray--plane lookup. Within this model, target projections and loop residuals change little after correction. This checkerboard procedure lets vision researchers examine LiDAR direction models alongside camera intrinsics and extrinsics, without a dedicated large calibration field.
\end{samepage}

\section*{LLM Usage Statement}
The authors used OpenAI ChatGPT/Codex and Anthropic Claude as tools to assist with literature search, exploration of experimental concepts and analysis of results, code development, figure preparation, and manuscript drafting and revision. Literature search results were checked against the cited sources. The authors evaluated the tools' suggestions, made the final scientific and editorial decisions, and take responsibility for the methods, results, references, and claims.

{\small\raggedright\setlength{\bibsep}{1pt plus .2pt}
}
\onecolumn
\phantomsection
\addcontentsline{toc}{section}{Supplementary Material}
\begin{center}{\Large\bfseries Supplementary Material}\end{center}
\setcounter{section}{0}\setcounter{figure}{0}\setcounter{table}{0}
\renewcommand{\thesection}{S\arabic{section}}
\renewcommand{\thefigure}{S\arabic{figure}}
\renewcommand{\thetable}{S\arabic{table}}
\renewcommand{\theHsection}{supp.\arabic{section}}
\renewcommand{\theHfigure}{supp.\arabic{figure}}
\renewcommand{\theHtable}{supp.\arabic{table}}
\captionsetup{hypcap=false}
\section{Evaluation references in prior calibration studies}\label{supp:literature}

Table~\ref{tab:literature_references} groups reported values by evaluation setting and identifies their reference and aggregation. Different references and statistics preclude a direct accuracy ranking.

\begin{table}[!htb]
\caption{Evaluation references and reported quantities in representative calibration studies. Translation values are in millimeters.}
\label{tab:literature_references}
\centering
\small
\setlength{\tabcolsep}{5pt}
\renewcommand{\arraystretch}{1.15}
\begin{tabular}{@{}>{\raggedright\arraybackslash}p{0.25\linewidth}>{\raggedright\arraybackslash}p{0.35\linewidth}>{\raggedright\arraybackslash}p{0.35\linewidth}@{}}
\toprule
Evaluation setting and reference & Study and configuration & Reported quantity \\
\midrule
Real rig: target-derived calibration & Koide et al.~\cite{koide2023}; sphere-target reference, multiple data pairs & Transform difference: 10\,mm and $0.132^\circ$ \\
\addlinespace
Real rig: composed camera transform & Fu et al.~\cite{fu2023}; stereo transform from Kalibr & Translation difference between composed LiDAR--camera transforms and stereo reference: 2--4\,mm \\
\addlinespace
Real rig: target registration & FAST-Calib~\cite{fastcalib2025}; circular-hole centers & Point-to-point registration residual: $\leq6.5$\,mm \\
\midrule
Real data, synthetic miscalibration & LCCNet~\cite{lccnet2021}; KITTI calibration reference, temporal filtering & Mean per-axis absolute translation error: 2.97\,mm; mean per-axis angular error: $0.017^\circ$ \\
\addlinespace
Simulated measurements and truth & Beltr\'an et al.~\cite{beltran2022}; monocular/HDL-64, three target poses & Mean Euclidean translation error: 8.2\,mm; mean rotation error: $2.4\times10^{-3}$\,rad \\
\midrule
Real rig: physical sensor-origin reference & This work; six rigs, 13 recordings, recording-excluded correction & Mean Euclidean translation-reference distance: 3.04\,mm \\
\bottomrule
\end{tabular}
\end{table}

\section{Reproducing the calibration}\label{supp:implementation}

Returns are preselected by nominal-ray intersection with the board interior under a fixed rigid transform. Tape-excluded intensities use channel-wise 5th--95th percentile scaling, clipped to $[0,1]$; fewer than 25 returns or a span below 45\% of the observation-wide span triggers observation-wide scaling. Valid observations place the board in front of the camera, have board reprojection RMS $\le1$\,px, and contain $\ge80$ inner-checker returns. A fixed list of 35--48 test observations per condition is used for all fits; the remaining valid observations form the fitting set.

Support is the longest run of $5\degree$ observation-mean elevation bins with $\ge50$ fitting observations in each of $\ge2$ recordings: L1 5--65$\degree$, L2/L3 10--70$\degree$, L4/L6 10--60$\degree$, L5 15--60$\degree$. Seven equally spaced knots have $\pm30$\,mrad bounds and constant end-knot extension. SLSQP uses analytic gradients and extrinsic bounds of $\pm50$\,mrad/$\pm100$\,mm around the rigid initialization. Starts use zero correction or per-recording calibration estimates. All 78 held-out solutions converge inside the bounds. Each evaluated recording is excluded from curve and support estimation; only its extrinsics are fitted. Recordings are equally weighted, without a spline roughness penalty.

Additional sensors use five knots bounded by $\pm60$\,mrad on the fixed four-recording grid; common supports are QT128 $[-16,24]\degree$, Airy $[13,57]\degree$, and JT128 $[16,56]\degree$. QT128 intensity rays rotate about native $+z$ by $\beta=\operatorname{atan2}(14.4\,\mathrm{mm},\sqrt{x^2+y^2})$, using native metric return coordinates and the transmitter--receiver spacing given in the QT128 manual. Parallax preprocessing uses range; subsequent lookup uses the fixed rays. Additional-sensor structure tests correct only elevation within common support.

\noindent\textbf{Timing.} Board returns span a median of 6.2\,ms (95th percentile 17.8\,ms). With the shared curves fixed, interpolating the synchronized board pose to each return's time changes the refitted extrinsics by a median of 0.31\,mm and $0.009\degree$ (maximum 0.91\,mm and $0.048\degree$).

\noindent\textbf{Code and data.} We plan to release code and data for reproducing the main results.

\section{Observation diversity and synthetic recovery}\label{supp:observability}

\noindent\textbf{Observation diversity.} When a checker is observed over a narrow region, an elevation adjustment can resemble a change in sensor pose. Broad azimuth coverage helps distinguish elevation offsets from rigid rotation; observations at multiple depths help distinguish angular changes from translation; and elevation coverage constrains the correction curve's shape.

For each of six units, we analyze all fitting observations from 13 recordings. We separately restrict azimuths to $\pm5\degree$, ray--board distances to $\pm10\%$, or elevations to $\pm5\degree$ around each recording's median. Random subsets match each restricted point count while retaining broad coverage, separating the effects of coverage and sample count.

\begin{figure}[!htb]
\centering
\includegraphics[width=0.86\linewidth]{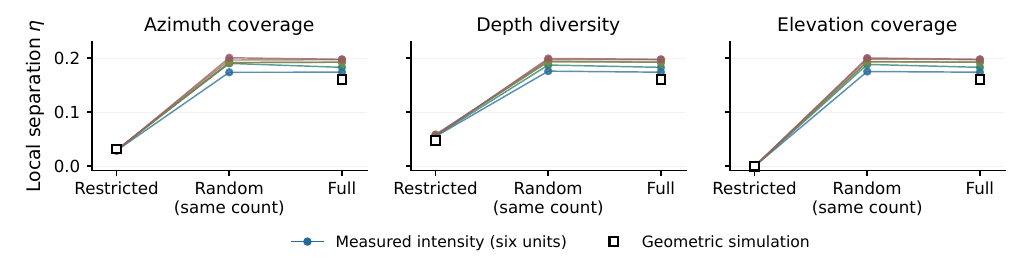}
\caption{Observation diversity and local separation. Colored lines: six units. Open squares: 4,096 simulated rays with geometric bearing residuals and a fixed spline basis, varying only azimuth half-width ($5\degree/40\degree$), depth (2\,m / 1, 1.5, 2.5, 4\,m), or elevation half-width ($5\degree/25\degree$, centered at $40\degree$). Larger $\eta$ means less local ambiguity, not greater absolute accuracy.}
\label{fig:coverage}
\end{figure}

We measure the fraction of the residual change caused by a small curve adjustment that remains after the best compensating extrinsic adjustment. Its minimum across curve adjustments, $\eta$, quantifies local separation: larger values mean that extrinsic changes are less able to mimic a curve change. Restricted coverage lowers $\eta$, whereas the random controls remain close to the full data (Fig.~\ref{fig:coverage}). Narrow elevation coverage also leaves a curve mode unobserved.

Centered, unit-norm measured and predicted intensities $p,q$ give the residual $r=p-(p^Tq)q$, with squared norm $1-\rho^2$. Let $J_c,J_e$ be its curve and extrinsic Jacobians with the fitting weights. Then $\eta=\min_{\delta\vect{c}\ne0}\|(I-J_eJ_e^+)J_c\delta\vect{c}\|/\|J_c\delta\vect{c}\|$, where $+$ denotes the pseudoinverse; an unobserved curve mode is assigned zero. This analysis is local to the fitted solution, with camera and board models fixed. For all six units, the full-data joint Jacobian has rank 85: seven curve coefficients and 13 six-parameter extrinsic blocks.

\noindent\textbf{Synthetic recovery.} Synthetic data use every fourth real fitting observation, retaining rays, poses and support. Truth extrinsics are the shared fits using all fitting observations; truth curves are fitted curves, their $\pm2$\,mrad shifts, or zero. Intensities are generated from a checker convolved with a Gaussian of $\sigma=11$\,mm plus independent Gaussian noise calibrated to the measured mean $\rho_k^2$. One noise realization is used per unit and case. Across six units, nonzero truth gives maximum absolute level error 0.24\,mrad and maximum unit-mean absolute $t_z$ error 0.43\,mm. Zero truth gives maxima of 0.15\,mrad and 0.33\,mm; shape RMS is $\le0.10$\,mrad (4--11\% of measured-shape RMS), with recovered shapes negatively correlated with measured shapes. On L1, slopes of $\pm0.1$\,mrad/$\degree$, 2\,mrad per-channel elevation or azimuth errors, and independent per-observation Gaussian board-pose perturbations at nominal and tripled standard deviations kept the level within 0.07\,mrad and mean absolute $t_z$ error within 0.2\,mm; a 0.1\% camera scale error moved $t_z$ by 0.08\,mm. For board depth $Z$ in meters, the nominal standard deviations were $0.1(Z/1.5)$\,mm laterally, $0.2(Z/1.5)^2$\,mm in depth, and $0.05(Z/1.85)/\sqrt{3}$ degrees per rotation-vector axis.

\section{Physical and camera-scan references}\label{supp:references}

\noindent\textbf{Physical reference.} The JT128 manual defines the data origin on the mechanical rotation axis, 47.84\,mm above the housing base (Sec. 1.3 and Fig. 5). Unit-specific factory angle corrections and the optical-center offset transformation (Appendix D) express decoded returns in this specified coordinate frame. Downstream extrinsic calibration estimates the pose of this same frame. Its documented origin is therefore held fixed as the reference when evaluating fitted translations. Arc-fixture camera centers fit a 229.9\,mm radius (RMS 0.11\,mm, maximum 0.18\,mm). CAD locates the center 5.4\,mm behind the lens front, giving $\mathbf{t}_{\mathrm{ref}}=[0,-94.76,-22.53]$\,mm. In two acquisitions with lens fronts aligned to a plane, removing a common row tilt gives depth RMS values of 0.13 and 0.17\,mm (maximum 0.30\,mm; RMS 0.12--0.28\,mm under alternative intrinsic solutions). Shifting the optical-axis component of this reference by $\pm2$\,mm leaves at least 71/78 conditions with a smaller full translation-reference distance after correction.

\noindent\textbf{Camera and mounting controls.} We compare relative camera translations across five recordings and bracket mount heights across rigs. The focal control propagates the observed 2.4\,px focal-length spread through board-pose estimation to $t_z$. Re-seating compares extrinsics before and after reinstalling the same LiDAR.

\noindent\textbf{Uncertainty intervals.} The 95\% intervals use 50,000 crossed-bootstrap draws, independently resampling units and recordings with factory--corrected pairing preserved, conditional on fitted models and reference.

\noindent\textbf{Camera-scan references.} RANSAC on 400,000 camera-scan points identifies planes, refined three times using points within 10\,mm. Each plane's 25\,cm spatial cells are resampled with replacement in 200 replicates, refitting planes and recomputing angles. Reported $\pm$ values are angle standard deviations ($1\sigma$), conditional on fixed intrinsics and reconstruction. Varying each reference within $\pm1\sigma$ leaves 147/180 JT128 pairs improving throughout, 28 worsening throughout, and five reference-dependent; Airy's 18/18 and QT128's 14/18 counts are unchanged.

\section{Calibration comparisons and consistency}\label{supp:comparisons}

All comparisons use the same inner-checker region, $x\in[0.01,0.39]$\,m and $y\in[0.01,0.63]$\,m, with at least 80 inner-checker returns per observation. Board poses and intrinsics are fixed. Corrected translation fits learn the shared curve from the other 12 recordings and refit only the excluded recording's extrinsics. Structure and image evaluations use the full 13-recording curve. Reported ranges are unchanged.

\noindent\textbf{Point-to-plane.} We minimize squared distances to camera-reconstructed board planes with equal observation weights. The joint variant adds the same data-origin seven-knot elevation model as the checker fits. This is a controlled extension of the plane constraint, not a full intrinsic model with fitted range parameters. The 78 recording-excluded source fits converge without active coefficient bounds; the minimum direction-mapping derivative is 0.868.

\noindent\textbf{Line and plane.} We apply Zhou et al.'s formulation~\cite{zhou2018}. Camera lines come from the known board outline and multiview poses instead of image-line detection. LiDAR boundaries use a cached extraction, a $15^\circ$ edge-direction quality check, and plane-inlier points within the checker region. The median accepted fitting-frame fraction is 43\%, with at least 186 frames per condition. All 78 conditions converge from six starts to the same translation at the reported precision.

\noindent\textbf{Checker formulations.} We implemented Fu et al.'s measured-point checkerboard formulation from the published equations, as no author-provided implementation was available to us. We evaluate standardized SSD and additionally introduce squared Pearson correlation and the shared seven-knot elevation model for controlled comparisons with ray--plane lookup. Phase-checked initializations fix checker polarity.

\noindent\textbf{Range perturbation.} We add 30\,mm to decoded ranges while holding post-preprocessing directions, origins, selected returns, intensities, board poses and initialization rules fixed, then refit only extrinsics. The main table reports mean translation change. The ray--plane fits reproduce their unperturbed solutions within $10^{-4}$\,mm. The line-and-plane comparison retains the selected boundary samples; it does not reselect views after perturbation.

\noindent\textbf{Projection displacement.} Each method's corrected projections are compared with its own factory-direction projections of the same held-out returns. Displacements are averaged within frames, then equally across frames within conditions and across conditions. For ray--plane lookup, 3,445 frames give mean displacement 0.96\,px, with condition means 0.50--1.84\,px; translation changes span 0.18--13.66\,mm. These quantities are changes, not ground-truth reprojection errors.

\noindent\textbf{Loop consistency.} Independent LiDAR-to-camera fits use single-camera board poses on even 2\,s blocks. Their relative transform is compared with an image-only stereo transform from odd blocks (median half-split repeatability 0.23\,mm). Two recordings, six units and camera pairs C1--C2 and C1--wideC1 provide 24 candidate loops. Joint rows use fixed corrections estimated without the evaluated recording. One L6 Fu-style fit to wideC1 fails the polarity check, so that loop is excluded from every method. All loop summaries use the remaining 23 conditions (Table~\ref{tab:benchmark_loop_pairs}). For ray--plane lookup, we average $\|\mathbf{t}_{\mathrm{corrected}}-\mathbf{t}_{\mathrm{factory}}\|$ over the two camera fits in each loop; its median over loops is 6.5\,mm. The median absolute paired change in loop residual is 0.003\,mm (maximum 0.02\,mm).

\begin{table}[!htb]
\centering
\caption{Loop translation residuals (mm) on the same 23 conditions for every method. One L6 loop using wideC1 is excluded from all methods after a Fu-style fit fails the polarity check. Pair columns report medians; Mean uses all 23 conditions.}
\label{tab:benchmark_loop_pairs}
\resizebox{.72\textwidth}{!}{%
\begin{tabular}{llrrrr}
\toprule
Method & Elev. & Mean & Median & C1--C2 & C1--wideC1 \\
\midrule
Line + plane & No & 2.06 & 1.17 & 0.65 & 3.59 \\
Point-to-plane & No & 5.00 & 2.24 & 1.59 & 10.67 \\
Point-to-plane & Yes & 4.99 & 2.25 & 1.60 & 10.53 \\
Fu-style + SSD & No & 8.71 & 6.95 & 3.88 & 11.50 \\
Fu-style + SSD & Yes & 9.33 & 7.12 & 3.32 & 13.15 \\
Fu-style + squared Pearson & No & 7.27 & 5.50 & 3.41 & 10.02 \\
Fu-style + squared Pearson & Yes & 8.03 & 5.62 & 2.74 & 11.90 \\
Ray + squared Pearson & No & 1.10 & 0.32 & 0.14 & 2.05 \\
Ray + squared Pearson & Yes & 1.10 & 0.31 & 0.14 & 2.05 \\
\bottomrule
\end{tabular}}
\end{table}

\section{Transfer across recordings and observing cameras}\label{supp:transfer}

\noindent\textbf{Acquisition-date exclusion.} We exclude each acquisition date when estimating the curve and its support (five dates and six units; 30 fits), then fit only extrinsics to fitting observations from that date. The held-out observations are re-filtered to the support learned from the remaining recordings. Across 78 unit--recording evaluations, the mean translation-reference distance decreases from 7.62\,mm with factory directions to 3.01\,mm with correction.

\noindent\textbf{Transfer across observing cameras.} We pair observations of the same board by each LiDAR and two cameras, matching timestamps, returns, normalization and elevation support. Targets are the six main cameras C1--C6; own-to-target camera-center separations span 0.09--0.86\,m. Tracking physical LiDAR units across the L2/L6 mounting exchange gives 32 sensor--target-camera groups and 387 paired recording conditions. Board poses are independently estimated from each camera's images with fixed intrinsics. The camera chain initializes extrinsics and expresses results in a common frame. The paired-camera protocol requires at least eight inner-checker returns per observation, rather than the main analysis's 80. Both lookup controls use squared Pearson correlation and a seven-knot spline.

Curve difference is the RMS over common observed elevation support. Replacing multi-view poses with own-camera PnP poses on these same observations gives a mean curve RMS difference of 0.16\,mrad (groups averaged within units, then across units). The 0.24\,mrad value instead compares observing cameras. For the other-camera translation metric, the own-camera curve is fixed and only the extrinsics are refitted. After transport through the fixed camera chain into the own-camera frame, we compare the LiDAR origin with the same physical reference:
\[
\|\Delta\mathbf{t}_{\mathrm{other}}\|
=\|\mathbf{t}_{\mathrm{other}\rightarrow\mathrm{own}}-\mathbf{t}_{\mathrm{ref}}\|.
\]
The reference is the manufacturer-defined LiDAR origin relative to the own camera's projection center. It is not the own-camera fitted translation. We average over recordings, camera groups and physical units in that order.

\begin{center}
\begin{minipage}{\textwidth}
\captionof{table}{Shared-curve changes and physical-reference translation differences on the same paired-camera observations. Own-camera values use this paired subset, rather than the main 78-condition cohort.}
\label{tab:camera_transfer}
\centering\small
\setlength{\tabcolsep}{12pt}
\begin{tabular}{lrrr}
\toprule
Lookup & Curve RMS difference & \multicolumn{2}{c}{Physical-reference difference (mm)} \\
 & (mrad) & Own camera & Other camera \\
\midrule
Ray--plane & 0.24 & 3.08 & 3.23 \\
Measured-point (Fu-based) & 1.12 & 3.92 & 10.49 \\
\bottomrule
\end{tabular}
\end{minipage}
\end{center}

Ray--plane curve differences are smaller in 29/32 groups. For the other constraints in main Table 5, line--plane boundaries support 381/387 pairs. Source bounds are reached in 13/32 factory and 23/32 corrected point-to-plane groups, and 4/32 corrected Fu-style SSD groups. Target bounds occur in two factory and one corrected point-to-plane fits and one corrected Fu-style Pearson fit. Factory and corrected Fu-style Pearson each retain one phase warning. These bounded-fit results use the fixed camera chain for physical-reference evaluation, not independent absolute ground truth.

Applying the ray--plane curves estimated with the own and other cameras to the separate structures gives mean inter-plane angle errors of $0.325^\circ$ and $0.324^\circ$, respectively. Both retain improvement over $0.453^\circ$ with factory directions.

\section{Structure and image-boundary evaluation}\label{supp:structures}

\noindent\textbf{LiDAR surface evaluation.} Stationary scans from twelve six-unit recordings provide 198 planes and 180 pairs. Orthogonal least squares fits identical points independently, without orthogonality constraints. Unit normals face the observed free space; inter-plane angle is $\theta=\pi-\arccos(\mathbf{n}_1^T\mathbf{n}_2)$ in degrees and error is $|\theta-\theta_{\mathrm{ref}}|$. Only elevation changes; ranges stay fixed and extrinsics are not used. Incidence-related range biases can remain~\cite{laconte2019}. End-knot values apply outside curve support (6.8\% of returns). Between-unit corner-angle spread does not decrease.

\noindent\textbf{Image-boundary measurements.} JT128 segment MAE and RMS measure perpendicular distances from projected samples to locally detected image-gradient lines. Segment scores, including RMS, are averaged equally with identical factory/corrected detector settings. Across 42 wall--floor segments, mean RMS falls from 5.36 to 3.25\,px; all 38 stable under detector changes improve (mean MAE $5.32\to3.30$\,px). QT128/Airy image MAE uses perpendicular distances from projected plane intersections to reference image boundaries, with model-matched extrinsics and identical retained samples.

\noindent\textbf{Additional-sensor preprocessing.} Vendor-driver re-decoding verified Airy's documented origin correction~\cite{robosenseairy}; the maximum channel-median elevation difference was $3.3\times10^{-5}$\,mrad. QT128 uses identical factory/corrected parallax preprocessing.

\clearpage
\section{Sensitivity to correction parameterization}\label{supp:parameterization}

The offset-plus-scale model fits $\alpha'=b+(1+s)\alpha$; splines use 7 or 13 equally spaced knots. Radial models use $\beta=\pi/2-\alpha$ from fixed native $+z$, $v=\beta/(\pi/2)$, and $\beta'=\beta+\sum_{j\in\{1,3,\ldots,d\}}a_jv^j$, with $f(\alpha)=\beta-\beta'$ and $d\in\{7,9\}$. Their four or five coefficients include linear scale. Total radial correction is bounded by $\pm30$\,mrad with monotonic direction mapping. Models share camera poses, checker samples, splits and support. Table~\ref{tab:parameterization_supp} evaluates full curves using fixed surfaces and image lines, model-matched extrinsics and constant endpoint extension. All source and target fits converge within bounds.

\begingroup
\setlength{\intextsep}{6pt plus 1pt minus 1pt}
\begin{table}[!htb]
\caption{Correction parameterizations on identical observations. Translation uses 78 recording-excluded fits; structure and image metrics use full curves on 198 planes, 180 pairs and 60 image segments. Shading marks the main model. Angles use the measured references.}
\label{tab:parameterization_supp}
\centering
\footnotesize
\setlength{\tabcolsep}{4pt}
\renewcommand{\arraystretch}{1.12}
\begin{tabular}{@{}lrrrrrrrr@{}}
\toprule
 & & \shortstack{Mean $\|\Delta\mathbf{t}\|$\\(mm)} & \multicolumn{3}{c}{Structure geometry} & \multicolumn{3}{c}{Image-edge MAE (px)} \\
\cmidrule(lr){4-6}\cmidrule(lr){7-9}
Elevation model & Coefficients & & \shortstack{Plane RMS\\(mm)} & \shortstack{Angle MAE\\($\degree$)} & \shortstack{Pairs\\improved} & Wall--floor & Wall--wall & Overall \\
\midrule
Factory directions & 0 & 7.57 & 6.734 & 0.453 & --- & 5.31 & 1.51 & 4.17 \\
Offset + scale & 2 & 3.10 & 6.695 & 0.424 & 118/180 & 3.51 & 1.74 & 2.98 \\
\rowcolor{inkblue!8}
Spline, 7 knots (used) & 7 & 3.04 & 6.713 & 0.343 & 151/180 & 3.22 & 1.58 & 2.73 \\
Spline, 13 knots & 13 & 2.98 & 6.795 & 0.349 & 150/180 & 3.08 & 1.56 & 2.62 \\
\midrule
Radial, degree 7 & 4 & 3.07 & 6.723 & 0.330 & 151/180 & 3.16 & 1.58 & 2.68 \\
Radial, degree 9 & 5 & 3.04 & 6.725 & 0.353 & 153/180 & 3.19 & 1.58 & 2.70 \\
\bottomrule
\end{tabular}
\end{table}
\endgroup

\end{document}